\documentclass[letterpaper, 10 pt, conference]{ieeeconf}  % Comment this line out if you need a4paper

\IEEEoverridecommandlockouts                              % This command is only needed if 
\usepackage{booktabs} % 用于处理美观的表格
\usepackage{graphicx} % 用于处理图像
\usepackage{amsmath} % 用于处理符号加粗
\usepackage{balance}  % 在导言区引入balance包
\usepackage{xcolor}    % 颜色支持
\usepackage{amssymb}   % \checkmark 等符号
\usepackage{multirow}

\newcommand{\cmark}{\textcolor{green}{\checkmark}}
\newcommand{\xmark}{\textcolor{red}{$\times$}}

\title{\LARGE \bf
Service Discovery-Based Hybrid Network Middleware for Efficient Communication in Distributed Robotic Systems
}

\author{Shiyao Sang$^{1,2}$ and Yinggang Ling$^{1}$% <-this % stops a space
\thanks{*This work was conducted at Huaiyin Institute of Technology and Chery Automobile Co., Ltd.}% <-this % stops a space
\thanks{$^{1}$Faculty of Computer and Software Engineering, Huaiyin Institute of Technology, Jiangsu, China.}%
\thanks{$^{2}$Chery Automobile Co., Ltd., Anhui, China.}%
\thanks{Corresponding author: {\tt\small www.fengmao@outlook.com}}%
}

\begin{document}

\maketitle
\thispagestyle{empty}
\pagestyle{empty}

%%%%%%%%%%%%%%%%%%%%%%%%%%%%%%%%%%%%%%%%%%%%%%%%%%%%%%%%%%%%%%%%%%%%%%%%%%%%%%%%
\begin{abstract}
Robotic middleware is fundamental to ensuring reliable communication among system components and is crucial for intelligent robotics, autonomous vehicles, and smart manufacturing. However, existing robotic middleware often struggles to meet the diverse communication demands, optimize data transmission efficiency, and maintain scheduling determinism between Orin computing units in large-scale L4 autonomous vehicle deployments. This paper presents RIMAOS2C, a service discovery-based hybrid network communication middleware designed to tackle these challenges. By leveraging multi-level service discovery multicast, RIMAOS2C supports a wide variety of communication modes, including multiple cross-chip Ethernet protocols and PCIe communication capabilities. The core mechanism of the middleware, the Message Bridge, optimizes data flow forwarding and employs shared memory for centralized message distribution, reducing message redundancy and minimizing transmission delay uncertainty, thus improving both communication efficiency and scheduling stability. Tested and validated on L4 vehicles and Jetson Orin domain controllers, RIMAOS2C leverages TCP-based ZeroMQ to overcome the large-message transmission bottleneck inherent in native CyberRT middleware. In scenarios involving two cross-chip subscribers, RIMAOS2C eliminates message redundancy, enhancing transmission efficiency by 36\%–40\% for large data transfers while reducing callback time differences by 42\%–906\%. This research advances the communication capabilities of robotic operating systems and introduces a novel approach to optimizing communication in distributed computing architectures for autonomous driving systems.

\end{abstract}

%%%%%%%%%%%%%%%%%%%%%%%%%%%%%%%%%%%%%%%%%%%%%%%%%%%%%%%%%%%%%%%%%%%%%%%%%%%%%%%%
\section{INTRODUCTION}

With the rapid development of artificial intelligence and robotics \cite{nilssonPrinciplesArtificialIntelligence1982, lecunDeepLearning2015}, distributed robotic operating systems (ROS) are increasingly being applied in fields such as intelligent robotics \cite{tsardouliasRoboticFrameworksArchitectures2017}, autonomous driving \cite{rekeSelfdrivingCarArchitecture2020a, belluardoMultiDomainSoftwareArchitecture2021}, and smart manufacturing \cite{sooriIntelligentRoboticSystems2024}. Unlike traditional embedded operating systems, robotic operating systems provide hardware abstraction and standardized software interfaces, significantly improving software portability and hardware compatibility \cite{elkadyRoboticsMiddlewareComprehensive2012a, liuComputingSystemsAutonomous2021}. This, in turn, facilitates the widespread deployment of robotic technologies across a wide range of industries.

\begin{figure}[t]
\vspace{5pt}
\centering
\includegraphics[width=3.4in]{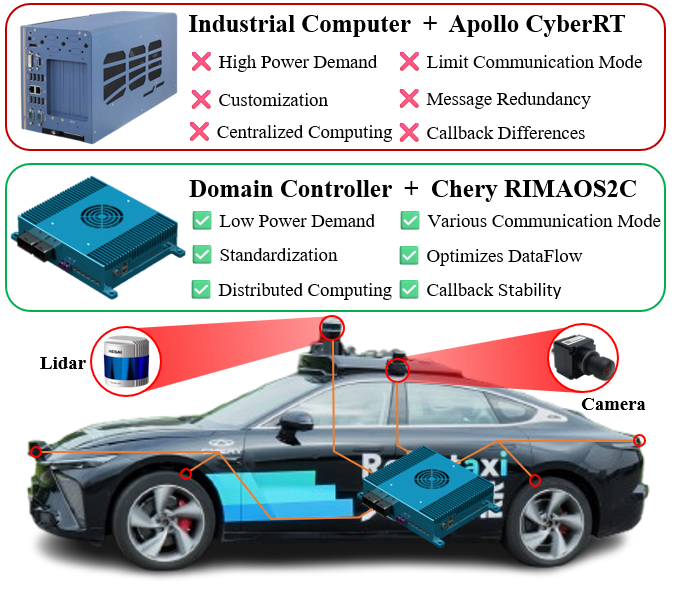}
\caption{Comparsion of Traditional Industrial Computer and Apollo CyberRT Solution\cite{ApolloAutoApolloOpen} vs. Proposed Domain Controller and Chery RIMAOS2C Solution}
\label{solution}
\vspace{-15pt}
\end{figure}

Robotic middleware serves as the central communication hub within robotic operating systems \cite{liuComputingSystemsAutonomous2021, smartCommonMiddlewareRobotics2007a}, facilitating high-level communication protocols, abstracting low-level complexities, and dynamically selecting the optimal communication mode based on node topology (e.g., intra-process, inter-process, or cross-chip communication). However, most existing robotic middleware is designed for general robotic systems, making them less suited for the complexities of distributed computing environments \cite{zhouSwarmMicroFlying2022}. These frameworks face significant limitations, including restricted communication modes \cite{zhouSwarmMicroFlying2022}, message redundancy \cite{maruyamaExploringPerformanceROS22016}, and scheduling uncertainties \cite{liuComputingSystemsAutonomous2021}.  

In the large-scale, standardized deployment of L4 autonomous vehicles using Jetson Orin domain controllers (see Fig. \ref{solution}), these limitations become increasingly evident when managing data exchange between Orin computing units. The existing service discovery-based automatic communication mode selection mechanism is limited to shared pointers, shared memory, and Data Distribution Service (DDS) \cite{innovationsOMGDataDistributionService} for network communication. However, distributed computing architectures for autonomous driving demand communication methods that provide low latency, high throughput, and efficient resource utilization for chip-to-chip data transmission \cite{zhouSwarmMicroFlying2022}. Furthermore, in multi-chip architectures, current middleware transmits identical data separately to multiple subscribers, leading to significant message redundancy, particularly when handling high-resolution sensor data such as LiDAR point clouds and camera images \cite{maruyamaExploringPerformanceROS22016}. Additionally, instability in inter-chip communication affects temporal consistency and processing stability, which in turn compromises data synchronization \cite{liuComputingSystemsAutonomous2021, saitoPrioritySynchronizationSupport2016}. These challenges degrade the real-time performance and safety of autonomous driving systems, highlighting the need for a more efficient, scalable, and adaptive middleware solution.

To address these challenges, this paper presents RIMAOS2C, a robotic middleware designed for distributed computing architectures in L4 autonomous driving systems. Utilizing a multi-level service discovery design, RIMAOS2C incorporates support for ZeroMQ \cite{ZeroMQ} and PCIe transmission via shared memory bridging, allowing automatic adaptation to various communication modes. Furthermore, RIMAOS2C introduces an efficient data flow forwarding mechanism that reduces message redundancy between distributed computing units. Additionally, RIMAOS2C employs a shared memory-based cross-chip message distribution mechanism, enhancing scheduling stability.

RIMAOS2C not only optimizes bandwidth utilization and minimizes communication overhead in L4 autonomous driving distributed computing architectures, but also ensures data transmission consistency and accelerates synchronous responses within autonomous driving computing systems. These advancements enhance the communication capabilities of the robotic middleware, while also boosting the real-time performance and safety of autonomous driving systems.

The main contributions of this work are as follows:
\begin{itemize}
\item We design and implement RIMAOS2C, a hybrid network communication middleware based on service discovery, which has been tested and validated on L4 autonomous driving vehicles in distributed computing architectures (see Fig. \ref{solution}).
\item RIMAOS2C employs a hybrid communication approach, using UDP-based DDS for small messages and TCP-based ZeroMQ for large messages, overcoming DDS's bottleneck in large message transmission.
\item RIMAOS2C incorporates a multi-level service discovery mechanism, resolving the challenge of integrating multiple communication modes and seamlessly embedding them into a unified robotic middleware framework.
\item RIMAOS2C utilizes a Message Bridge for data flow management and a shared memory-based centralized distribution mechanism to resolve issues of message redundancy and callback inconsistencies caused by communication fluctuations.
\end{itemize}

\section{Related Work}

As robotic systems are increasingly deployed in complex environments, the demands on their communication mechanisms have grown significantly. Robotic middleware plays a crucial role as a bridge for seamless interaction between system components, facilitating data exchange, task scheduling, and resource management within distributed computing environments \cite{elkadyRoboticsMiddlewareComprehensive2012a, jungOpenSourceAutonomousDriving2025}. Widely adopted frameworks like DDS, Robot Operating System (ROS1/ROS2) \cite{quigleyROSOpensourceRobot2009, macenskiRobotOperatingSystem2022}, and CyberRT \cite{IntroductionCyberRT} have greatly enhanced development efficiency in areas such as industrial automation, autonomous driving, and distributed robotics. These frameworks have significantly advanced the deployment of distributed robotic systems.

A key feature of modern robotic middleware is its ability to automatically discover communication endpoints and select appropriate communication methods based on system configuration and network topology \cite{teixeiraServiceOrientedMiddleware2011}. Traditional ROS1 relies on a centralized master node for service discovery, which introduces a single point of failure and is unsuitable for large-scale distributed computing environments. In contrast, ROS2 and CyberRT employ a decentralized UDP multicast service discovery mechanism \cite{ROSDDS, TopologicalDiscoveryCommunication}, enhancing scalability in distributed systems.

DDS is the cross-chip communication method used by ROS2 and CyberRT \cite{ROSDDS, vanderperkDistributedSafetyMechanism2019}. While DDS provides Quality of Service (QoS) adjustments to optimize communication across different scenarios \cite{maruyamaExploringPerformanceROS22016, cruzDDSbasedMiddlewareQualityofservice2012}, its reliance on UDP-based data transmission introduces significant performance bottlenecks in high-throughput environments \cite{jungOpenSourceAutonomousDriving2025, 154LargeData}. Additionally, current frameworks do not support alternative communication modes, such as TCP-based transmission optimizations or PCIe direct connections, which could substantially improve communication efficiency between computational units.

Recent studies have explored optimizing communication protocols to enhance the performance of robotic middleware. For instance, the QUIC protocol \cite{langleyQUICTransportProtocol2017}, which combines the low-latency benefits of UDP with the reliability of TCP, improves network transmission stability. SCTP \cite{eklundUsingMultiplePaths2018}, with its multi-stream transmission and advanced congestion control mechanisms, significantly reduces network delay and mitigates the impact of scheduling fluctuations. Additionally, Software-Defined Networking (SDN) \cite{kreutzSoftwaredefinedNetworkingComprehensive2014, zhangSDNFVFlexibleDynamic2016} and Edge Computing \cite{bonomiFogComputingIts2012} are leveraged for dynamic bandwidth management and localized data processing, reducing network transmission load. However, these approaches are largely application-specific and have not been integrated into a unified robotic middleware architecture. As a result, they leave significant gaps in addressing the need for diverse and efficient communication methods, message redundancy, and scheduling uncertainty—issues that are especially critical in large-scale L4 computing architectures.

\section{SYSTEM DESIGN}
\subsection{Distributed Computing Architecture}
\begin{figure*}
\centering
\includegraphics[width=6.8in]{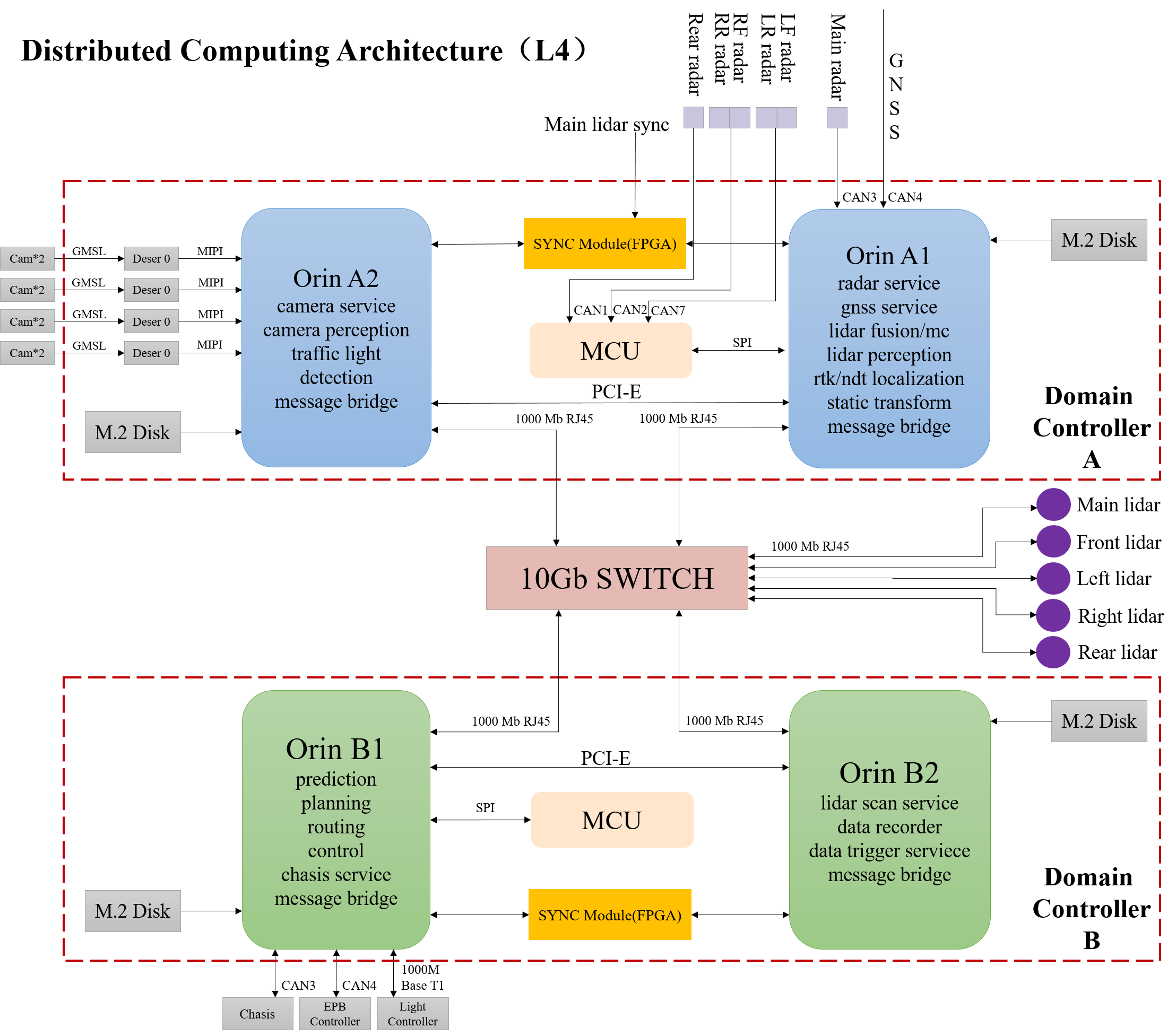}
\caption{The L4 autonomous driving distributed computing architecture, integrated with the RIMAOS2C robotic middleware, provides flexible deployment and computational scalability. This architecture employs four Orin computing units working in unison to execute key functions, including LiDAR perception, visual perception, localization, planning, control, and data acquisition.}
\label{hardware}
\vspace{-15pt}
\end{figure*}

The distributed computing architecture for autonomous driving consists of two domain controllers, each containing two Orin chips and one microcontroller unit (MCU). These Orin computing units are interconnected via a 10Gb Ethernet switch to enable efficient data exchange. Within each domain controller, the two Orin chips communicate through a PCIe-based virtual network, facilitating low-latency, high-bandwidth data sharing for computational tasks.

As shown in Fig. \ref{hardware}, the functional allocation across the domain controllers is structured as follows. Domain Controller A is primarily responsible for perception tasks: Orin A1 processes LiDAR and millimeter-wave radar sensing, sensor data fusion, and localization, while Orin A2 focuses on visual perception, including object detection and traffic light recognition. Domain Controller B manages planning, control, and data collection functions: Orin B1 handles prediction, planning, control, and chassis computations, whereas Orin B2 is dedicated to data collection and storage, ensuring reliable data recording. Additionally, the allocation of these modules can be dynamically adjusted across chips based on system load and hardware configuration.

Although this distributed architecture offers high modularity and scalability, it also presents new communication challenges. The following section will outline the core design and implementation of RIMAOS2C.

\subsection{RIMAOS2C Middleware Design}

\begin{figure}[t]
\vspace{5pt}
\centering
\includegraphics[width=3.2in]{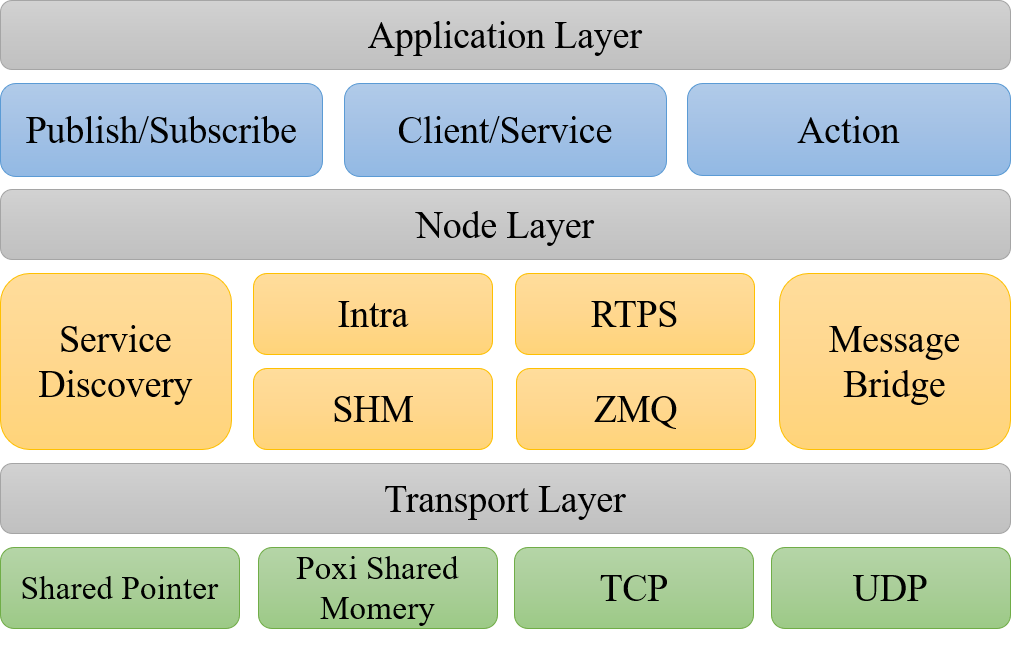}
\caption{RIMAOS2C Software Architecture Diagram}
\label{framework}
\vspace{-10pt}
\end{figure}

\begin{figure}
\centering
\includegraphics[width=3.2in]{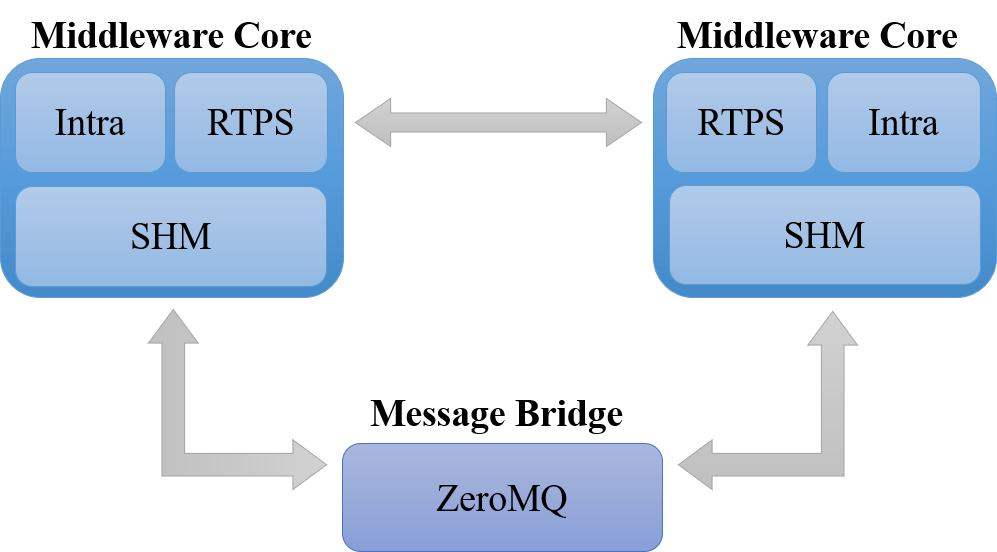}
\caption{RIMAOS2C Communication Design Diagram}
\label{communication}
\vspace{-15pt}
\end{figure}

RIMAOS2C is a robotic middleware solution built on a node network, extending the core principles of ROS-based middleware to meet the demands of complex distributed systems. As shown in Fig. \ref{framework}, RIMAOS2C is structured into three layers: the Application Layer, Node Layer, and Transport Layer, forming a flexible and efficient communication framework.

The Application Layer provides standard communication interfaces, including Publish/Subscribe, Client/Service, and Action modes, allowing developers to focus on business logic without managing the underlying communication details.

The Node Layer manages service discovery, node management, and message routing. Service discovery allows nodes to automatically select the most suitable communication method based on network topology and system requirements. Node management monitors the status of each node and maintains the network topology. Message routing enables dynamic adaptation to various communication modes, including intra-process, inter-process, and cross-chip communication. Additionally, RIMAOS2C incorporates a Message Bridge to optimize data flow, reduce redundancy, and minimize callback fluctuations.

The Transport Layer implements communication protocols and supports multiple modes, including Shared Pointer (Intra) for low-latency data exchange within the same process, POSIX-based Shared Memory (SHM) for efficient inter-process communication within the same chip, UDP-based DDS-RTPS for low-latency communication between chips, and TCP-based ZeroMQ (ZMQ) for high-throughput, large-message transmission between chips. Real-Time Publish-Subscribe (RTPS) is a wire protocol used in DDS implementations.

As shown in Fig. \ref{communication}, RIMAOS2C dynamically switches between shared pointers, shared memory, and DDS-RTPS based on service discovery and the physical relationships between nodes. Additionally, ZeroMQ is integrated into shared memory via the Message Bridge, enabling seamless adaptation to multiple communication modes. By combining service discovery with the Message Bridge, RIMAOS2C ensures efficient and flexible communication that evolves with the system's needs. The design challenges associated with this approach are discussed in the following section.

\subsection{Service Discovery}

\begin{figure}[t]
\vspace{2pt}
\centering
\includegraphics[width=3.2in]{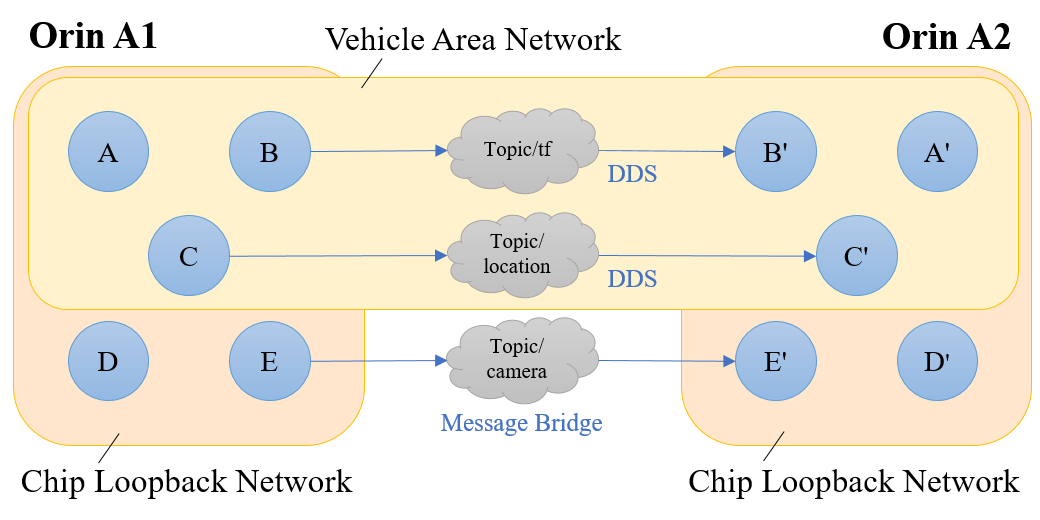}
\caption{Multi-level Service Discovery Diagram. Small messages (e.g., localization data) and large messages (e.g., image data) are separated into different discovery networks, enabling the use of distinct communication methods within the framework without encountering topic conflicts caused by automatic service discovery.}
\label{discovery}
\vspace{-20pt}
\end{figure}

Service discovery is a core functionality of robotic middleware, enabling the dynamic identification and interaction of distributed nodes. In RIMAOS2C, the service discovery mechanism inherits the design from ROS2 and CyberRT, utilizing a UDP multicast mechanism for automatic node discovery and status synchronization. Continuous state broadcasting and updates enhance the system's adaptability and robustness, ensuring real-time consistency across distributed nodes.

In ROS2 and CyberRT, the service discovery mechanism operates within a single discovery network, managing topic publication and subscription across the vehicle area network. However, when using a Message Bridge, duplicate topics from other Orin computing units may be introduced, potentially causing communication conflicts. While renaming topics can resolve these conflicts, it compromises the flexibility of distributed system deployment.

To resolve topic name conflicts, RIMAOS2C adopts multicast network isolation in its service discovery mechanism. As shown in Fig. \ref{discovery}, it employs a multi-level strategy, dividing networks into chip-local (127.0.0.1) and vehicle-area (192.168.1.x) segments. This design isolates communication scopes and avoids conflicts caused by automatic service discovery and additional Message Bridge communication across multiple Orin units. Chip-local discovery handles intra-chip communication, while vehicle-area discovery manages inter-chip communication.

This multi-level service discovery approach offers several advantages. By isolating networks, it minimizes service discovery conflicts while maintaining the flexibility of distributed deployment. It also ensures compatibility with existing CyberRT tools (e.g., cyber\_monitor and cyber\_recorder) \cite{ApolloCyberRT}, allowing users to monitor or record topics across different network segments. Furthermore, this approach is highly scalable and can be applied to more complex network topologies and diverse functional modules in distributed systems. Through multi-level service discovery isolation, RIMAOS2C effectively addresses DDS service discovery issues in distributed computing architectures, providing flexible deployment options for distributed robotic systems while maintaining compatibility with existing toolchains.

\subsection{Message Bridge}

\begin{figure}[t]
\vspace{2pt}
\centering
\includegraphics[width=3.2in]{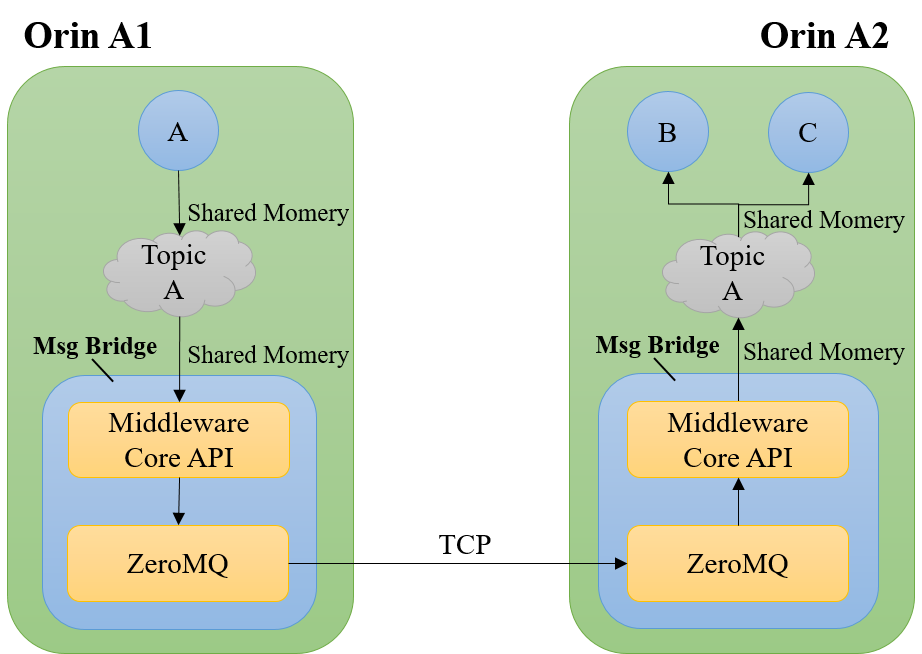}
\caption{Implementation Details of Message Bridge}
\label{msgbridge}
\vspace{-15pt}
\end{figure}

To bridge the differences between various communication protocols, RIMAOS2C introduces the Message Bridge, a specialized node that facilitates communication between different chips. The Message Bridge subscribes to messages from nodes via shared memory on the local chip, transmits these messages over the network to another chip, and republishes them into shared memory on the target chip (as shown in Fig. \ref{msgbridge}). This design transforms RIMAOS2C from a purely distributed communication system into a shared-memory-based, centralized communication architecture. Shared memory not only serves as the method for data transfer but also acts as a critical bridge for data exchange between Orin units.

The decision of which messages to forward through the Message Bridge is based on static configuration. However, the communication modes—shared pointers, shared memory, and DDS—are automatically adapted without requiring any static configuration. Additionally, the Message Bridge incorporates the ZeroMQ protocol, which utilizes both physical network channels and PCIe virtual network channels. PCIe communication can be up to four times faster than the physical network channel, enabling more efficient data transfer, particularly for tasks such as transmitting raw image messages.

This design improves distributed communication performance in several ways. The Message Bridge reduces message redundancy between Orin units by centrally forwarding data requested by multiple subscribers, thereby lowering network load and resource consumption. Additionally, the shared memory-based centralized distribution mechanism stabilizes communication by mitigating scheduling uncertainties between Orin units, enhancing synchronization response times and real-time performance in complex scenarios. However, this approach introduces a slight latency overhead (see Evaluation Section A).

The introduction of ZeroMQ in RIMAOS2C addresses the limitations of CyberRT's Fast DDS \cite{EProsimaFastDDS}, particularly in handling large-scale data transfers. While Fast DDS performs well for small messages, it struggles with large data volumes, often leading to instability or transmission failure. Modifying Fast DDS to support larger data transfers by adjusting its buffer size and QoS configurations would undermine its advantages in lightweight transmission \cite{jungOpenSourceAutonomousDriving2025, 154LargeData}. Therefore, RIMAOS2C adopts ZeroMQ, which is optimized for high throughput via TCP and PCIe virtual network channels, enabling more efficient large-data transmission. Simultaneously, it retains Fast DDS's low latency and flexible QoS configurations.

To further optimize ZeroMQ, RIMAOS2C limits its use to point-to-point communication between Orin units, addressing the `slow participant' issue \cite{ZGuideChapter1} often encountered in distributed systems. It establishes multiple data channels based on different topics and socket ports, enabling differentiated communication according to data size and priority. This approach allows for more refined transmission scheduling, significantly enhancing overall communication efficiency.

\section{EVALUATION}

To assess the communication efficiency and stability of RIMAOS2C in distributed computing architectures, a series of experiments were conducted. These tests focused on evaluating latency across different communication modes, cross-chip communication efficiency, and message scheduling stability. The experiments were conducted on Jetson Orin autonomous driving domain controllers to validate RIMAOS2C's optimizations and benchmark its performance against CyberRT. However, RIMAOS2C is designed to be cross-platform compatible and not limited to this hardware environment. ROS1 and ROS2 were excluded from the comparison due to the modifications required in their serialization mechanisms to ensure compatibility with autonomous driving applications—without which serialization design would significantly affect latency estimates \cite{jungOpenSourceAutonomousDriving2025}. In China, CyberRT has become the predominant middleware system for L4 autonomous vehicles, largely due to the widespread adoption of Baidu's Apollo \cite{ApolloAutoApolloOpen} open-source autonomous driving platform. Derived from ROS1, CyberRT represents an evolved middleware framework tailored to the specific requirements of robotic and autonomous driving systems in China.

\subsection{Latency Evaluation Across Various Communication Modes}

This experiment assesses the transmission latency of RIMAOS2C under various communication modes and verifies its adaptive strategy for different data scales. The modes tested include shared memory (SHM), Fast DDS (UDP), and ZeroMQ (TCP), with performance evaluated across different data sizes.

RIMAOS2C optimizes service discovery and utilizes the Message Bridge to centrally distribute messages using shared memory within devices. For cross-chip communication, Fast DDS (UDP) is employed for small data to minimize latency, while ZeroMQ (TCP) is used for larger data to enhance throughput. Since ZeroMQ transmits data through the Message Bridge, its measured latency includes additional overhead from two shared memory (SHM) operations. To account for the actual network transmission time, the latency for ZeroMQ is adjusted as follows:

\[
T_{\text{ZeroMQ (Adjusted)}} = T_{\text{ZeroMQ (Original)}} - 2 \times T_{\text{SHM}}
\]

where \( T_{\text{ZeroMQ (Original)}} \) represents the original latency of ZeroMQ, and \( T_{\text{SHM}} \) refers to the time spent on each shared memory operation. This adjustment provides a more accurate reflection of the network transmission time.

\begin{table}[ht]
    \vspace{5pt}
    \centering
    \caption{Comparison of Inter-Process Communication Latency (in $\mu$s)}
    \renewcommand{\arraystretch}{1.1} % 适当调整行距
    \begin{tabular}{p{1.1cm}cccccc}
        \toprule
        \multirow{2}{*}{\textbf{Method}} & \multicolumn{6}{c}{\textbf{Message Size}} \\
        \cmidrule(lr){2-7}
        & \textbf{1KB} & \textbf{10KB} & \textbf{100KB} & \textbf{1MB} & \textbf{2MB} & \textbf{6MB} \\
        \midrule
        \scriptsize SHM & 181 & 210 & 375 & 991 & 1779 & 4399 \\
        \scriptsize Fast DDS & 1340 & 1310 & 1948 & 11023 & 21635 & 65299 \\
        \scriptsize ZeroMQ \newline (Original) & 2182 & 2427 & 3223 & 12783 & 23341 & 66475 \\
        \scriptsize ZeroMQ \newline (Adjusted) & 1820 & 2007 & 2473 & 10701 & 19783 & 57677 \\
        \bottomrule
    \end{tabular}
    \label{table:ipc_latency}
    \vspace{-15pt}
\end{table}

Table \ref{table:ipc_latency} compares the inter-process communication latency (in microseconds, $\mu$s) across various communication methods: Shared Memory (SHM), Fast DDS, and ZeroMQ (Original and Adjusted). The data sizes tested range from 1KB to 6MB. Shared Memory (SHM) consistently delivers the lowest latency, as it operates within the memory space without involving network protocol stacks, minimizing communication delays. For small data sizes ($\leq$100KB), Fast DDS (UDP) outperforms ZeroMQ (TCP), showing lower latency for these smaller transmissions. However, for larger data sizes ($\geq$2MB), ZeroMQ (Adjusted) offers superior throughput, with significantly lower transmission times compared to Fast DDS (UDP). Fast DDS, on the other hand, struggles with large data, showing instability and frame loss when handling data over 2MB. To maintain stability, transmission frequency was reduced to one message per second, making Fast DDS unsuitable for large data volumes typically found in autonomous driving applications.

While RIMAOS2C also offers a non-Message Bridge version of ZeroMQ as a potential replacement for Fast DDS, it would require static configuration for all communication messages, conflicting with the design goal of dynamic communication adaptation based on service discovery.

The Message Bridge's use of shared memory reduces redundant data transmissions and enhances scheduling consistency. While this introduces additional read/write delays, the tradeoff is justified by the overall system performance improvements, leading to more efficient and stable communication.

\subsection{Performance Comparison: RIMAOS2C vs. CyberRT}
These experiments compare the latency, efficiency, and scheduling consistency of RIMAOS2C and CyberRT across various communication modes. The test scenarios encompass communication latency between Orin computing units, data transmission efficiency for multiple subscribers across Orin units, and callback consistency for multiple subscribers.

In the experimental setup, CyberRT relies mainly on DDS (UDP Unicast) for communication, while RIMAOS2C utilizes DDS (UDP Unicast) and the Message Bridge to optimize large data transmission via ZeroMQ (TCP). RIMAOS2C also introduces a centralized shared memory distribution mechanism, enhancing callback consistency.

\subsubsection{Latency Comparison}

The first test focused on comparing the communication latency between RIMAOS2C and CyberRT under various data sizes. Due to RIMAOS2C's use of adaptive switching between shared memory, DDS, and ZeroMQ, its latency performance depends on the chosen communication mode.

\begin{figure}[t]
\vspace{5pt}
\centering
\includegraphics[width=3.2in]{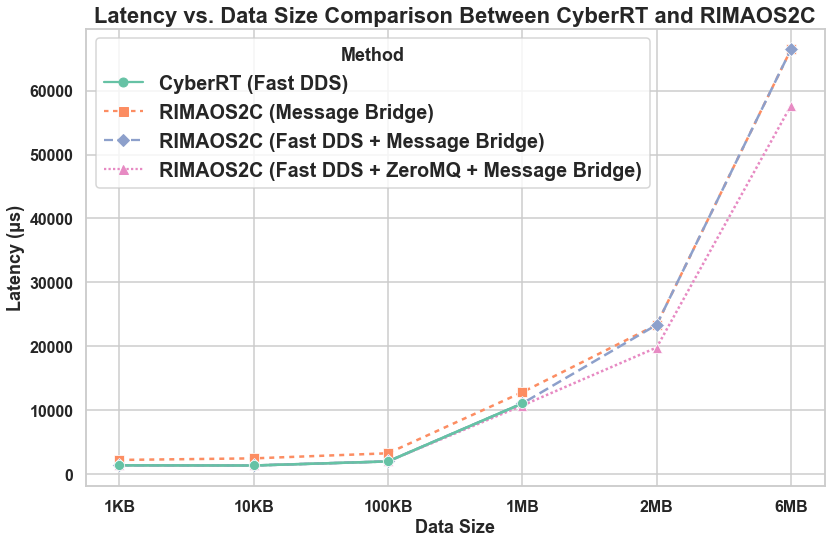}
\caption{Compared to CyberRT, which only supports Fast DDS, RIMAOS2C enables multiple communication methods, allowing for an optimized combination to achieve the best communication performance.}
\label{comdiffconfig}
\vspace{-15pt}
\end{figure}

As shown in Fig. \ref{comdiffconfig}, in small data scenarios ($\leq$100KB), CyberRT (Fast DDS) exhibits lower latency, leveraging the efficiency of the UDP protocol for small data transmission. In contrast, for larger data sizes ($\geq$2MB), RIMAOS2C (Message Bridge) ensures stable transmission, overcoming CyberRT's bottleneck in handling large message communication between Orin units. By integrating different communication methods, RIMAOS2C optimizes performance across various data loads.

\subsubsection{Inner-Orin Message Transmission Efficiency Comparison}

This experiment evaluates message redundancy in RIMAOS2C and CyberRT during multi-subscriber cross-chip communication, with a focus on transmission efficiency, which directly impacts network bandwidth utilization and latency. The test simulates multiple subscribers receiving the same topic and analyzes the performance of the Message Bridge-optimized ZeroMQ transmission in reducing redundancy and improving efficiency.
\begin{itemize}
\item Sub1 and Sub2 subscribe via Fast DDS (UDP Unicast).
\item Sub3 and Sub4 subscribe via Message Bridge (ZeroMQ + SHM).
\end{itemize}

\begin{figure}[t]
\vspace{5pt}
\centering
\includegraphics[width=3.2in]{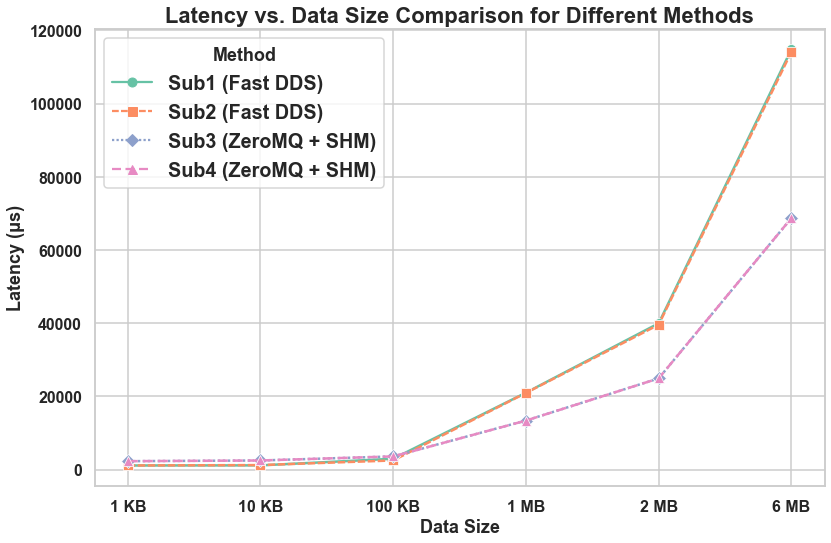}
\caption{As the message size increases, the advantages of RIMAOS2C's data forwarding mechanism become more apparent, providing more stable transmission.}
\label{comeff}
\vspace{-15pt}
\end{figure}

As shown in Fig. \ref{comeff}, RIMAOS2C reduces communication latency by 36\% to 40\% when handling data larger than 100KB, thanks to the optimization provided by the Message Bridge. The primary cause of this improvement lies in the fact that the Message Bridge transmits cross-chip communication only once. This strategy reduces transmission latency, improves transmission efficiency, and decreases network overhead, thereby increasing the effective bandwidth available for communication.

The reduction in latency can be quantified using the following formula:

\[
\Delta L = \frac{(L_{\text{Fast DDS}} - L_{\text{ZeroMQ + SHM}})}{L_{\text{Fast DDS}}} \times 100\%
\]

where \( L_{\text{Fast DDS}} \) represents the communication latency when using Fast DDS in CyberRT, and \( L_{\text{ZeroMQ + SHM}} \) corresponds to the latency achieved with the ZeroMQ + SHM method. This calculation demonstrates the efficiency gains from adopting RIMAOS2C's optimized communication strategy.

\subsubsection{Scheduling Stability Across Different Load Conditions}

This experiment assesses the callback time difference in RIMAOS2C under various data sizes to evaluate its scheduling consistency. Similar to the previous experiment, it compares the performance of Message Bridge (Bridge) and Fast DDS (Unicast) subscriptions by measuring the callback time differences between two subscribers.

\begin{table}[ht]
    \centering
    \footnotesize
    \caption{Average Difference in Latency (in $\mu$s)}
    \renewcommand{\arraystretch}{1.1} % 适当调整行距
    \begin{tabular}{p{1.2cm}cccccc}
        \toprule
        \multirow{2}{*}{\textbf{Middleware}} & \multicolumn{6}{c}{\textbf{Message Size}} \\
        \cmidrule(lr){2-7}
         & \textbf{1KB} & \textbf{10KB} & \textbf{100KB} & \textbf{1MB} & \textbf{2MB} & \textbf{6MB} \\
        \midrule
        \scriptsize CyberRT \cite{vanderperkDistributedSafetyMechanism2019} \newline (Fast DDS) & 92.48 & 116.02 & 467.89 & 255.69 & 746.68 & 773.81 \\
        \scriptsize RIMAOS2C \newline (ZeroMQ + SHM) & \textbf{54.09} & \textbf{55.00} & \textbf{53.47} & \textbf{70.95} & \textbf{68.28} & \textbf{86.48} \\
        \bottomrule
    \end{tabular}
    \label{table:avg_diff_latency}
    \vspace{-5pt}
\end{table}

As shown in Table \ref{table:avg_diff_latency}, RIMAOS2C reduces the callback time difference by 42\% to 906\%, significantly enhancing scheduling stability. This improvement plays a crucial role in accelerating data synchronization response times. In contrast, CyberRT demonstrates greater fluctuations in callback times under high-load conditions, highlighting its limited ability to maintain stable scheduling performance during heavy communication loads.

The reduction in callback time difference can be quantified as follows:

\[
\Delta T = \frac{(T_{\text{CyberRT}} - T_{\text{RIMAOS2C}})}{T_{\text{CyberRT}}} \times 100\%
\]

where \( T_{\text{CyberRT}} \) denotes the callback time different in CyberRT, and \( T_{\text{RIMAOS2C}} \) represents the callback time different in RIMAOS2C. This formula illustrates the effectiveness of RIMAOS2C in minimizing callback time differences.

By mitigating fluctuations in callback times, RIMAOS2C ensures more consistent data synchronization, which is vital for improving overall scheduling stability. This becomes especially important in high-load environments, such as those encountered in autonomous driving systems, where timely and reliable synchronization is essential for both safety and performance.

\subsection{Feature Comparison: RIMAOS2C and Other Middleware Systems}

\begin{table*} % 使用 table* 环境实现跨双栏
    \vspace{5pt}
    \centering
    \footnotesize % 使用较小字体，避免内容超出页面宽度
    \caption{Comparison of Different Robotics Middleware}
    \label{tab:middleware_comparison}
    \begin{tabular}{lccccccccc}
    \toprule
    \textbf{Middleware} & \shortstack[t]{\textbf{Service}\\\textbf{Discovery}} & \textbf{SHM} & \textbf{UDP (QoS)} & \textbf{TCP} & \textbf{PCIe} & \shortstack[t]{\textbf{Dataflow}\\\textbf{Forwarding}} & \shortstack[t]{\textbf{SHM}\\\textbf{Distribution}} & \shortstack[t]{\textbf{Distributed}\\\textbf{Capability}} \\
    \midrule
    ROS1 \cite{quigleyROSOpensourceRobot2009}    & Master Node      & \xmark & \xmark & \cmark & \xmark & \xmark  & \xmark & Medium \\
    ROS2 \cite{macenskiRobotOperatingSystem2022}   & Single Multicast & \cmark & \cmark & \xmark & \xmark & \xmark & \xmark & High \\
    CyberRT \cite{IntroductionCyberRT}                   & Single Multicast & \cmark & \cmark & \xmark & \xmark & \xmark  & \xmark & High \\
    RIMAOS2C                                      & Multiple Multicast & \cmark & \cmark & \cmark & \cmark & \cmark  & \cmark & Medium \\
    \bottomrule
    \end{tabular}
    \vspace{-5pt}
\end{table*}

Table \ref{tab:middleware_comparison} presents a comparative analysis of RIMAOS2C, ROS1, ROS2, and CyberRT, emphasizing key differences in communication mechanisms, service discovery, and distributed capability. Unlike ROS1, which relies on a centralized Master Node, and ROS2/CyberRT, which utilize single-multicast DDS-based discovery, RIMAOS2C implements multi-multicast service discovery, improving communication flexibility and adaptability in distributed architectures. While ROS1 lacks SHM support and relies primarily on TCP, ROS2 and CyberRT incorporate SHM and UDP with QoS but lack native support for TCP and PCIe, limiting their efficiency in large-scale data transmission. RIMAOS2C extends communication capabilities by integrating TCP (ZeroMQ) and PCIe, enhancing throughput and reducing latency in cross-chip data exchange. Furthermore, RIMAOS2C introduces a Message Bridge with a data flow forwarding mechanism, effectively reducing message redundancy and optimizing bandwidth utilization. Although its shared memory-based distribution mechanism slightly reduces its distributed capability compared to the fully decentralized architectures of ROS2 and CyberRT, this trade-off results in improved scheduling stability and lower callback latency fluctuations. These advancements make RIMAOS2C particularly well-suited for real-time applications such as L4 autonomous driving, where efficient cross-chip communication and low-latency message processing are critical.

\section{CONCLUSION AND FUTURE WORK}

In this study, we designed and implemented RIMAOS2C, a robotic middleware system that employs a service discovery-based hybrid network communication mechanism to address key challenges in existing robotic middleware. RIMAOS2C utilizes multi-level service discovery via multicast, ensuring compatibility with various communication methods within a unified framework. It reduces message redundancy and mitigates scheduling uncertainty by employing efficient data flow forwarding and a shared memory-based distribution mechanism. Experimental results show that RIMAOS2C outperforms CyberRT in high-load data scenarios ($\geq$2MB), overcoming CyberRT's cross-chip data bottleneck. In multi-subscriber scenarios, RIMAOS2C transmits cross-chip data only once, optimizing bandwidth and improving scheduling stability under heavy loads. These features make RIMAOS2C well-suited for real-time applications such as autonomous driving and distributed robotics.

Despite its demonstrated improvements, the Message Bridge forwarding mechanism in RIMAOS2C inherently compromises the peer-to-peer nature typical of distributed systems, potentially limiting its applicability in multi-agent robotic scenarios. Future work will focus on developing role-based or context-aware dynamic activation strategies for the bridge, allowing selective activation based on factors such as task priority, node computational load, and communication context. These strategies aim to balance the trade-off between communication efficiency and system decentralization. Additionally, integrating RIMAOS2C into a unified, high-level communication framework capable of dynamically adapting transmission modes according to message size, latency constraints, and priority is envisioned. This would enable more efficient network resource utilization, reduce redundant transmissions, and improve scheduling determinism, particularly in complex or high-interference network environments.


%%%%%%%%%%%%%%%%%%%%%%%%%%%%%%%%%%%%%%%%%%%%%%%%%%%%%%%%%%%%%%%%%%%%%%%%%%%%%%%%

%%%%%%%%%%%%%%%%%%%%%%%%%%%%%%%%%%%%%%%%%%%%%%%%%%%%%%%%%%%%%%%%%%%%%%%%%%%%%%%%

%%%%%%%%%%%%%%%%%%%%%%%%%%%%%%%%%%%%%%%%%%%%%%%%%%%%%%%%%%%%%%%%%%%%%%%%%%%%%%%%

% \section*{ACKNOWLEDGMENT}

%%%%%%%%%%%%%%%%%%%%%%%%%%%%%%%%%%%%%%%%%%%%%%%%%%%%%%%%%%%%%%%%%%%%%%%%%%%%%%%%

\balance  % 在参考文献之前调用

\bibliographystyle{IEEEtran}
\bibliography{reference}%第二个参数参数是你的bib文件的名字

@inproceedings{quigleyROSOpensourceRobot2009,
  title = {{{ROS}}: An Open-Source {{Robot Operating System}}},
  shorttitle = {{{ROS}}},
  booktitle = {{{ICRA}} Workshop on Open Source Software},
  author = {Quigley, Morgan and Conley, Ken and Gerkey, Brian and Faust, Josh and Foote, Tully and Leibs, Jeremy and Wheeler, Rob and Ng, Andrew Y.},
  year = {2009},
  volume = {3},
  pages = {5},
  publisher = {Kobe, Japan},
  urldate = {2024-06-30}
}

@inproceedings{belluardoMultiDomainSoftwareArchitecture2021,
  title = {A {{Multi-Domain Software Architecture}} for {{Safe}} and {{Secure Autonomous Driving}}},
  booktitle = {2021 {{IEEE}} 27th {{International Conference}} on {{Embedded}} and {{Real-Time Computing Systems}} and {{Applications}} ({{RTCSA}})},
  author = {Belluardo, Luca and Stevanato, Andrea and Casini, Daniel and Cicero, Giorgiomaria and Biondi, Alessandro and Buttazzo, Giorgio},
  year = {2021},
  month = aug,
  pages = {73--82},
  publisher = {IEEE},
  address = {Houston, TX, USA},
  doi = {10.1109/RTCSA52859.2021.00017},
  urldate = {2024-06-30},
  isbn = {978-1-66544-188-9}
}

@inproceedings{maruyamaExploringPerformanceROS22016,
  title = {Exploring the Performance of {{ROS2}}},
  booktitle = {Proceedings of the 13th {{International Conference}} on {{Embedded Software}}},
  author = {Maruyama, Yuya and Kato, Shinpei and Azumi, Takuya},
  year = {2016},
  month = oct,
  pages = {1--10},
  publisher = {ACM},
  address = {Pittsburgh Pennsylvania},
  doi = {10.1145/2968478.2968502},
  urldate = {2024-06-30},
  isbn = {978-1-4503-4485-2},
  langid = {english}
}

@article{kreutzSoftwaredefinedNetworkingComprehensive2014,
  title = {Software-Defined Networking: {{A}} Comprehensive Survey},
  shorttitle = {Software-Defined Networking},
  author = {Kreutz, Diego and Ramos, Fernando MV and Verissimo, Paulo Esteves and Rothenberg, Christian Esteve and Azodolmolky, Siamak and Uhlig, Steve},
  year = {2014},
  journal = {Proceedings of the IEEE},
  volume = {103},
  number = {1},
  pages = {14--76},
  publisher = {Ieee},
  urldate = {2024-06-30}
}

@inproceedings{bonomiFogComputingIts2012,
  title = {Fog Computing and Its Role in the Internet of Things},
  booktitle = {Proceedings of the First Edition of the {{MCC}} Workshop on {{Mobile}} Cloud Computing},
  author = {Bonomi, Flavio and Milito, Rodolfo and Zhu, Jiang and Addepalli, Sateesh},
  year = {2012},
  month = aug,
  pages = {13--16},
  publisher = {ACM},
  address = {Helsinki Finland},
  doi = {10.1145/2342509.2342513},
  urldate = {2024-06-30},
  isbn = {978-1-4503-1519-7},
  langid = {english}
}

@inproceedings{zhangSDNFVFlexibleDynamic2016,
  title = {{{SDNFV}}: {{Flexible}} and {{Dynamic Software Defined Control}} of an {{Application-}} and {{Flow-Aware Data Plane}}},
  shorttitle = {{{SDNFV}}},
  booktitle = {Proceedings of the 17th {{International Middleware Conference}}},
  author = {Zhang, Wei and Liu, Guyue and Mohammadkhan, Ali and Hwang, Jinho and Ramakrishnan, K. K. and Wood, Timothy},
  year = {2016},
  month = nov,
  pages = {1--12},
  publisher = {ACM},
  address = {Trento Italy},
  doi = {10.1145/2988336.2988338},
  urldate = {2024-06-30},
  isbn = {978-1-4503-4300-8},
  langid = {english}
}

@inproceedings{langleyQUICTransportProtocol2017,
  title = {The {{QUIC Transport Protocol}}: {{Design}} and {{Internet-Scale Deployment}}},
  shorttitle = {The {{QUIC Transport Protocol}}},
  booktitle = {Proceedings of the {{Conference}} of the {{ACM Special Interest Group}} on {{Data Communication}}},
  author = {Langley, Adam and Riddoch, Alistair and Wilk, Alyssa and Vicente, Antonio and Krasic, Charles and Zhang, Dan and Yang, Fan and Kouranov, Fedor and Swett, Ian and Iyengar, Janardhan and Bailey, Jeff and Dorfman, Jeremy and Roskind, Jim and Kulik, Joanna and Westin, Patrik and Tenneti, Raman and Shade, Robbie and Hamilton, Ryan and Vasiliev, Victor and Chang, Wan-Teh and Shi, Zhongyi},
  year = {2017},
  month = aug,
  pages = {183--196},
  publisher = {ACM},
  address = {Los Angeles CA USA},
  doi = {10.1145/3098822.3098842},
  urldate = {2024-06-30},
  isbn = {978-1-4503-4653-5},
  langid = {english}
}

@article{innovationsOMGDataDistributionService,
  title = {{{OMG Data-Distribution Service}}: {{Architectural Overview}}},
  shorttitle = {{{OMG Data-Distribution Service}}},
  author = {Innovations, Real-Time},
  urldate = {2024-06-30}
}

@misc{ROSDDS,
  title = {{{ROS}} on {{DDS}}},
  urldate = {2024-06-30},
  howpublished = {https://design.ros2.org/articles/ros\_on\_dds.html}
}

@book{nilssonPrinciplesArtificialIntelligence1982,
  title = {Principles of Artificial Intelligence},
  author = {Nilsson, Nils J.},
  year = {1982},
  publisher = {Springer Science \& Business Media},
  urldate = {2024-06-30}
}

@article{lecunDeepLearning2015,
  title = {Deep Learning},
  author = {LeCun, Yann and Bengio, Yoshua and Hinton, Geoffrey},
  year = {2015},
  journal = {nature},
  volume = {521},
  number = {7553},
  pages = {436--444},
  publisher = {Nature Publishing Group UK London},
  urldate = {2024-06-30}
}

@misc{IntroductionCyberRT,
  title = {Introduction --- {{Cyber RT Documents}} Documentation},
  urldate = {2024-06-30},
  howpublished = {https://cyber-rt.readthedocs.io/en/latest/index.html}
}

@article{macenskiRobotOperatingSystem2022,
  title = {Robot {{Operating System}} 2: {{Design}}, Architecture, and Uses in the Wild},
  shorttitle = {Robot {{Operating System}} 2},
  author = {Macenski, Steven and Foote, Tully and Gerkey, Brian and Lalancette, Chris and Woodall, William},
  year = {2022},
  month = may,
  journal = {Science Robotics},
  publisher = {American Association for the Advancement of Science},
  doi = {10.1126/scirobotics.abm6074},
  urldate = {2024-06-30},
  copyright = {Copyright {\copyright} 2022 The Authors, some rights reserved; exclusive licensee American Association for the Advancement of Science. No claim to original U.S. Government Works},
  langid = {english}
}

@misc{TopologicalDiscoveryCommunication,
  title = {Topological {{Discovery}} and {{Communication Negotiation}}},
  urldate = {2024-08-01},
  howpublished = {https://design.ros2.org/articles/discovery\_and\_negotiation.html}
}

@misc{ZeroMQ,
  title = {{{ZeroMQ}}},
  urldate = {2024-08-01},
  howpublished = {https://zeromq.org/}
}

@article{elkadyRoboticsMiddlewareComprehensive2012a,
  title = {Robotics {{Middleware}}: {{A Comprehensive Literature Survey}} and {{Attribute-Based Bibliography}}},
  shorttitle = {Robotics {{Middleware}}},
  author = {Elkady, Ayssam and Sobh, Tarek},
  year = {2012},
  journal = {Journal of Robotics},
  volume = {2012},
  number = {1},
  pages = {959013},
  issn = {1687-9619},
  doi = {10.1155/2012/959013},
  urldate = {2024-08-11},
  copyright = {Copyright {\copyright} 2012 Ayssam Elkady and Tarek Sobh.},
  langid = {english}
}

@inproceedings{smartCommonMiddlewareRobotics2007a,
  title = {Is a Common Middleware for Robotics Possible?},
  booktitle = {Proceedings of the {{IROS}} 2007 Workshop on {{Measures}} and {{Procedures}} for the {{Evaluation}} of {{Robot Architectures}} and {{Middleware}}},
  author = {Smart, William D.},
  year = {2007},
  publisher = {Citeseer},
  urldate = {2024-08-11}
}

@article{cruzDDSbasedMiddlewareQualityofservice2012,
  title = {A {{DDS-based}} Middleware for Quality-of-Service and High-Performance Networked Robotics},
  author = {Cruz, Jes{\'u}s Mart{\'i}nez and {Romero-Garc{\'e}s}, Adri{\'a}n and Rubio, Juan Pedro Bandera and Robles, Rebeca Marfil and Rubio, Antonio Bandera},
  year = {2012},
  journal = {Concurrency and Computation: Practice and Experience},
  volume = {24},
  number = {16},
  pages = {1940--1952},
  issn = {1532-0634},
  doi = {10.1002/cpe.2816},
  urldate = {2024-08-11},
  copyright = {Copyright {\copyright} 2012 John Wiley \& Sons, Ltd.},
  langid = {english}
}

@inproceedings{rekeSelfdrivingCarArchitecture2020a,
  title = {A Self-Driving Car Architecture in {{ROS2}}},
  booktitle = {2020 {{International SAUPEC}}/{{RobMech}}/{{PRASA Conference}}},
  author = {Reke, Michael and Peter, Daniel and {Schulte-Tigges}, Joschua and Schiffer, Stefan and Ferrein, Alexander and Walter, Thomas and Matheis, Dominik},
  year = {2020},
  pages = {1--6},
  publisher = {IEEE},
  urldate = {2024-08-11}
}

@misc{tsardouliasRoboticFrameworksArchitectures2017,
  title = {Robotic Frameworks, Architectures and Middleware Comparison},
  author = {Tsardoulias, Emmanouil and Mitkas, Pericles},
  year = {2017},
  month = nov,
  number = {arXiv:1711.06842},
  eprint = {1711.06842},
  primaryclass = {cs},
  publisher = {arXiv},
  doi = {10.48550/arXiv.1711.06842},
  urldate = {2024-08-11},
  archiveprefix = {arXiv}
}

@article{sooriIntelligentRoboticSystems2024,
  title = {Intelligent Robotic Systems in {{Industry}} 4.0: {{A}} Review},
  shorttitle = {Intelligent Robotic Systems in {{Industry}} 4.0},
  author = {Soori, Mohsen and Dastres, Roza and Arezoo, Behrooz and Karimi Ghaleh Jough, Fooad},
  year = {2024},
  journal = {Journal of Advanced Manufacturing Science and Technology},
  pages = {2024007--0},
  doi = {10.51393/j.jamst.2024007},
  urldate = {2024-08-11}
}

@article{eklundUsingMultiplePaths2018,
  title = {Using Multiple Paths in {{SCTP}} to Reduce Latency for Signaling Traffic},
  author = {Eklund, Johan and Grinnemo, Karl-Johan and Brunstrom, Anna},
  year = {2018},
  month = sep,
  journal = {Computer Communications},
  volume = {129},
  pages = {184--196},
  issn = {0140-3664},
  doi = {10.1016/j.comcom.2018.07.016},
  urldate = {2024-08-12}
}

@inproceedings{saitoPrioritySynchronizationSupport2016,
  title = {Priority and {{Synchronization Support}} for {{ROS}}},
  booktitle = {2016 {{IEEE}} 4th {{International Conference}} on {{Cyber-Physical Systems}}, {{Networks}}, and {{Applications}} ({{CPSNA}})},
  author = {Saito, Yukihiro and Azumi, Takuya and Kato, Shinpei and Nishio, Nobuhiko},
  year = {2016},
  month = oct,
  pages = {77--82},
  publisher = {IEEE},
  address = {Nagoya, Japan},
  doi = {10.1109/CPSNA.2016.24},
  urldate = {2024-08-12},
  isbn = {978-1-5090-4403-0},
  langid = {english}
}

@misc{154LargeData,
  title = {15.4. {{Large Data Rates}} --- {{Fast DDS}} 3.1.2 Documentation},
  urldate = {2025-02-17},
  howpublished = {https://fast-dds.docs.eprosima.com/en/latest/fastdds/use\_cases/\allowbreak large\_data/large\_data.html}
}

@misc{ZGuideChapter1,
  title = {{{ZGuide Chapter}} 1 - {{Basics}} \# {{Getting}} the {{Message Out}}},
  urldate = {2025-02-17},
  howpublished = {https://zguide.zeromq.org/docs/chapter1/\#Getting-the-Message-Out},
  langid = {english}
}

@misc{ApolloCyberRT,
  title = {Apollo {{Cyber RT Developer Tools}} --- {{Cyber RT Documents}} Documentation},
  urldate = {2025-02-17},
  howpublished = {https://cyber-rt.readthedocs.io/en/latest/\allowbreak CyberRT\_Developer\_Tools.html}
}

@misc{ApolloAutoApolloOpen,
  title = {{{ApolloAuto}}/Apollo: {{An}} Open Autonomous Driving Platform},
  urldate = {2025-02-19},
  howpublished = {https://github.com/ApolloAuto/apollo}
}

@article{zhouSwarmMicroFlying2022,
  title = {Swarm of Micro Flying Robots in the Wild},
  author = {Zhou, Xin and Wen, Xiangyong and Wang, Zhepei and Gao, Yuman and Li, Haojia and Wang, Qianhao and Yang, Tiankai and Lu, Haojian and Cao, Yanjun and Xu, Chao and Gao, Fei},
  year = {2022},
  month = may,
  journal = {Science Robotics},
  publisher = {American Association for the Advancement of Science},
  doi = {10.1126/scirobotics.abm5954},
  urldate = {2025-02-27},
  copyright = {Copyright {\copyright} 2022 The Authors, some rights reserved; exclusive licensee American Association for the Advancement of Science. No claim to original U.S. Government Works},
  langid = {english}
}

@misc{jungOpenSourceAutonomousDriving2025,
  title = {Open-{{Source Autonomous Driving Software Platforms}}: {{Comparison}} of {{Autoware}} and {{Apollo}}},
  shorttitle = {Open-{{Source Autonomous Driving Software Platforms}}},
  author = {Jung, Hee-Yang and Paek, Dong-Hee and Kong, Seung-Hyun},
  year = {2025},
  month = jan,
  number = {arXiv:2501.18942},
  eprint = {2501.18942},
  primaryclass = {cs},
  publisher = {arXiv},
  doi = {10.48550/arXiv.2501.18942},
  urldate = {2025-02-27},
  archiveprefix = {arXiv}
}

@article{liuComputingSystemsAutonomous2021,
  title = {Computing {{Systems}} for {{Autonomous Driving}}: {{State}} of the {{Art}} and {{Challenges}}},
  shorttitle = {Computing {{Systems}} for {{Autonomous Driving}}},
  author = {Liu, Liangkai and Lu, Sidi and Zhong, Ren and Wu, Baofu and Yao, Yongtao and Zhang, Qingyang and Shi, Weisong},
  year = {2021},
  month = apr,
  journal = {IEEE Internet of Things Journal},
  volume = {8},
  number = {8},
  pages = {6469--6486},
  issn = {2327-4662},
  doi = {10.1109/JIOT.2020.3043716},
  urldate = {2025-02-28}
}

@inproceedings{teixeiraServiceOrientedMiddleware2011,
  title = {Service {{Oriented Middleware}} for the {{Internet}} of {{Things}}: {{A Perspective}}},
  shorttitle = {Service {{Oriented Middleware}} for the {{Internet}} of {{Things}}},
  booktitle = {Towards a {{Service-Based Internet}}},
  author = {Teixeira, Thiago and Hachem, Sara and Issarny, Val{\'e}rie and Georgantas, Nikolaos},
  editor = {Abramowicz, Witold and Llorente, Ignacio M. and Surridge, Mike and Zisman, Andrea and Vayssi{\`e}re, Julien},
  year = {2011},
  pages = {220--229},
  publisher = {Springer},
  address = {Berlin, Heidelberg},
  doi = {10.1007/978-3-642-24755-2_21},
  isbn = {978-3-642-24755-2},
  langid = {english}
}

@article{vanderperkDistributedSafetyMechanism2019,
  title = {A Distributed Safety Mechanism for Autonomous Vehicle Software Using Hypervisors},
  author = {{van der Perk}, Peter},
  year = {2019},
  journal = {MS thesis},
  publisher = {Dept. Elect. Eng., Eindhoven University of Technology},
  urldate = {2025-02-28}
}

@misc{EProsimaFastDDS,
  title = {{{eProsima Fast DDS Documentation}} --- {{Fast DDS}} 3.1.2 Documentation},
  urldate = {2025-03-01},
  howpublished = {https://fast-dds.docs.eprosima.com/en/latest/}
}

\end{document}